%% file: paper.tex
\documentclass[10pt,twocolumn,letterpaper]{article}

\usepackage[pagenumbers]{cvpr} 

\usepackage{graphicx}
\usepackage{amsmath}
\usepackage{amssymb}
\usepackage{booktabs}

\usepackage[pagebackref,breaklinks,colorlinks]{hyperref}

\usepackage[capitalize]{cleveref}
\crefname{section}{Sec.}{Secs.}
\Crefname{section}{Section}{Sections}
\Crefname{table}{Table}{Tables}
\crefname{table}{Tab.}{Tabs.}

\usepackage{color}
\usepackage{multirow}
\usepackage{lipsum}
\definecolor{gray}{RGB}{128,128,128}
\SetLipsumParListSurrounders{\begingroup\color{gray}}{\endgroup}

\begin{document}

\title{ProSFDA: Prompt Learning based Source-free Domain Adaptation for Medical Image Segmentation}

\author{Shishuai Hu, Zehui Liao, and Yong Xia$^{\dagger}$\\
School of Computer Science and Engineering, Northwestern Polytechnical University, China\\
{\tt\small \{sshu, merrical\}@mail.nwpu.edu.cn; yxia@nwpu.edu.cn}
}
\maketitle

\input{main.tex}


{\small
\bibliographystyle{ieee_fullname}
\bibliography{egbib}
}

\clearpage

\appendix
\setcounter{figure}{0}
\renewcommand{\thefigure}{A\arabic{figure}}
\input{supplementary.tex}

\end{document}

%% file: main.tex

\begin{abstract}
The domain discrepancy existed between medical images acquired in different situations renders a major hurdle in deploying pre-trained medical image segmentation models for clinical use. 
Since it is less possible to distribute training data with the pre-trained model due to the huge data size and privacy concern, source-free unsupervised domain adaptation (SFDA) has recently been increasingly studied based on either pseudo labels or prior knowledge.
However, the image features and probability maps used by pseudo label-based SFDA and the consistent prior assumption and the prior prediction network used by prior-guided SFDA may become less reliable when the domain discrepancy is large.
In this paper, we propose a \textbf{Pro}mpt learning based \textbf{SFDA} (\textbf{ProSFDA}) method for medical image segmentation, which aims to improve the quality of domain adaption by minimizing explicitly the domain discrepancy.
Specifically, in the prompt learning stage, we estimate source-domain images via adding a domain-aware prompt to target-domain images, then optimize the prompt via minimizing the statistic alignment loss, and thereby prompt the source model to generate reliable predictions on (altered) target-domain images.
In the feature alignment stage, we also align the features of target-domain images and their styles-augmented counterparts to optimize the source model, and hence push the model to extract compact features.
We evaluate our ProSFDA on two multi-domain medical image segmentation benchmarks.
Our results indicate that the proposed ProSFDA outperforms substantially other SFDA methods and is even comparable to UDA methods.
Code will be available at \url{https://github.com/ShishuaiHu/ProSFDA}.
\end{abstract}

\section{Introduction}

\begin{figure}[]
  \centering
  \includegraphics[scale=0.4]{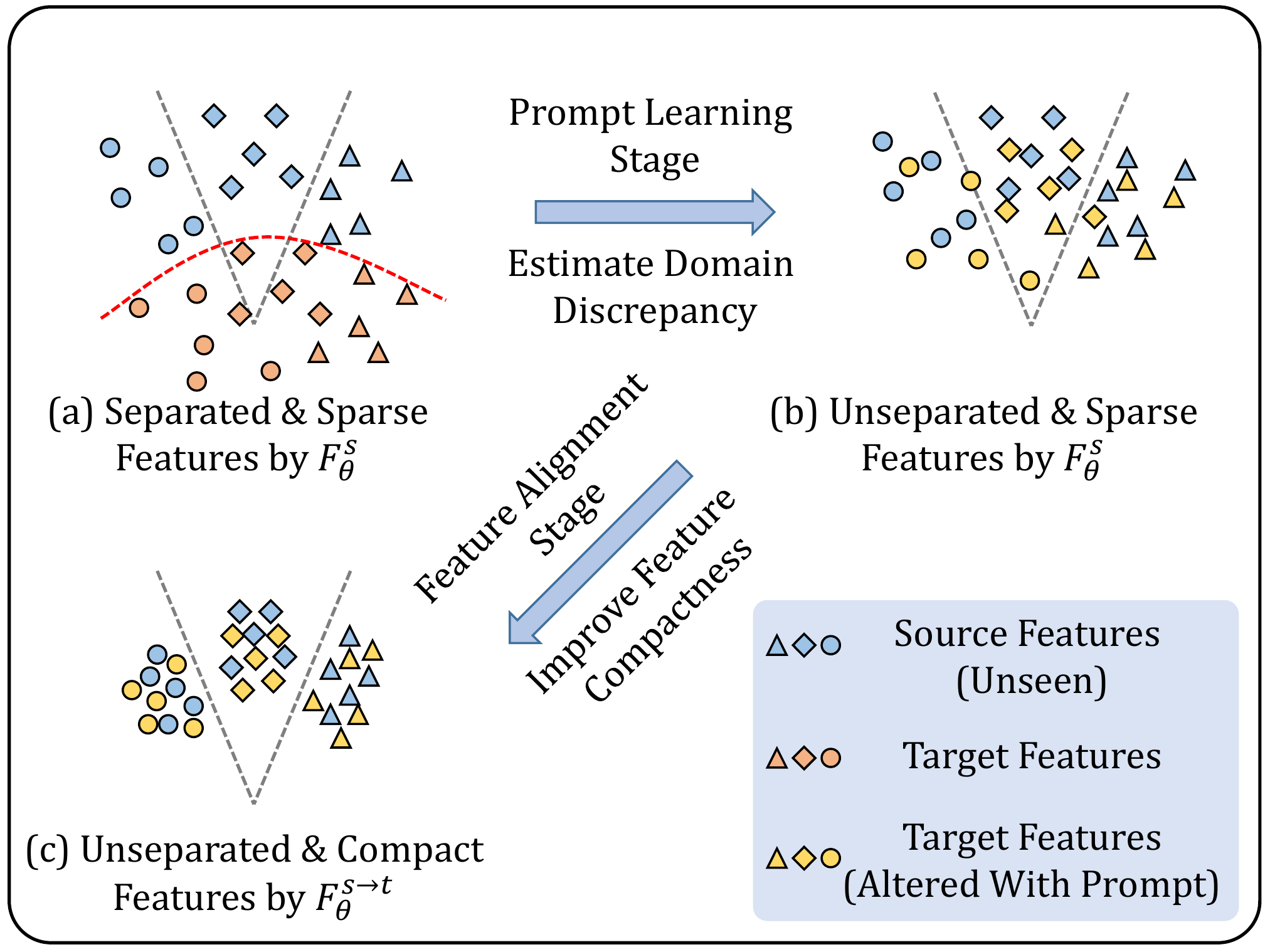}
  \caption{
Conceptual illustration of our ProSFDA. $F_{\theta}^s$: source model. $F_{\theta}^{s\to t}$: adapted model. The features in different shapes denote different classes. The dashed lines in gray and red represent class and domain boundaries respectively. 
Our ProSFDA estimates the domain discrepancy in the Prompt Learning Stage, and improves the compactness of features extracted from target images in the Feature Alignment Stage.
  }
  \label{fig:motivation}
\end{figure}

Deep learning models have achieved remarkable success in medical image segmentation~\cite{isensee_nnu-net_2020,falk_u-net_2019,LITJENS201760}.
Like any supervised learning approaches, these models are designed for the closed-world setting rooted in the assumption that training and test data are independent and identically distributed~\cite{guan2021domain,xie_survey_2021}. However, this assumption rarely holds in practice, as the distribution discrepancy commonly exists among different medical image datasets. This can be attributed to the fact that the range of image quality and structure visibility can be considerable, depending on the characteristics of the imaging equipment, the skill of the operator, and compromises with factors such as patient radiation exposure and imaging time~\cite{IMGCHAR}.
Consequently, when deploying a well-trained medical image segmentation model to a new and unseen dataset, unpredictable performance degradation becomes inevitable. 
To relieve this issue, unsupervised domain adaptation (UDA), where both labeled source-domain data and unlabeled target-domain data are assumed to be accessible during model training, has been extensively explored ~\cite{hu_domain_2022,yang_fda_2020,wang2019boundary,wilson2020survey}.
However, this assumption can hardly be held, since it is not practical to distribute the training data with the trained model, considering the huge data size and privacy concerns~\cite{bateson_source-relaxed_2020,bateson_source-free_2021}.
Compared to UDA, a more practical setting is source-free domain adaptation (SFDA)~\cite{liang_we_2020}, under which only the segmentation model pre-trained on labeled source-domain data (\ie, source model) is distributed to a client, and the model is then fine-tuned with unlabeled target-domain data.

Most SFDA methods target at mining the knowledge of absent source-domain data from the pre-trained source model and then transferring the knowledge to unlabeled target-domain data~\cite{yang_source_2022,liu_source-free_2022,liu_source-free_2021, liu_adapting_2021,eastwood_source-free_2022}.
An intuitive solution is to use the source model to generate pseudo labels for target-domain data and thus convert the problem into a supervised one on the target domain ~\cite{chen_source-free_2021,xu_denoising_2022,dong_confident_2021,chu_denoised_2022,ding_source-free_2022}.
Unavoidably, the pseudo labels contain a lot of noise due to the existence of the domain discrepancy. Hence, it is critical for pseudo label-based methods to identify and correct noisy labels, which remains challenging.
Besides, another drawback of these methods is that their performance depends heavily on the quality of the features and probability maps produced by the source model, which are considerable unreliable when the domain discrepancy is large.
To relieve the issues caused by pseudo labels, prior-guided SFDA methods~\cite{bateson_source-relaxed_2020,bateson_source-free_2021} have been proposed.
These methods minimize a label-free entropy loss over unlabeled target-domain data and use an additional prior knowledge prediction network to facilitate the knowledge transfer.
Although they avoid the use of pseudo labels during model adaptation, 
the reliability of their prior knowledge predictor is still influenced by the domain discrepancy.
To address the domain discrepancy issue, we advocate using prompt learning~\cite{jia_visual_2022,bahng_visual_2022,zhang_amortized_2022,zheng_prompt_2022} that enables us to estimate the discrepancy between source-domain data and target-domain data using only the pre-trained source model and unlabeled target-domain images.

In this paper, we propose a \textbf{Pro}mpt learning based \textbf{SFDA} method, called \textbf{ProSFDA}, for medical image segmentation.
The workflow of ProSFDA consists of a Prompt Learning Stage (PLS) and a Feature Alignment Stage (FAS).
In PLS, we add a domain-aware prompt, which is a learnable visual image, to each target-domain image and push the altered target-domain images to approximate source-domain ones via tuning the domain-aware prompt (see~\figurename{~\ref{fig:motivation}} (a)$\to$(b)).
Specifically, we first update the source model using the altered target-domain images.
Since it has been recognized that the statistics stored in batch normalization (BN) layers (\ie, BN statistics) are able to represent the domain information ~\cite{chang2019domain,liu2020ms,mancini2019inferring}, we then define the difference of BN statistics in the source model and updated source model as the statistic alignment loss, which measures the domain discrepancy.
Next, we optimize the domain-aware prompt via minimizing the statistic alignment loss.
Once the domain-aware prompt is optimized, it can represent the domain discrepancy, and the altered target-domain images can be used interchangeably with source-domain images (see~\figurename{~\ref{fig:motivation}} (b)).
However, due to the existence of instance variations, the feature distribution of target-domain images can be scattered, making it less possible to match the decision boundaries of the source model (see~\figurename{~\ref{fig:motivation}} (b)).
To improve the compactness of the feature distribution, we introduce FAS to align the feature of each altered target-domain image and the feature of its style-augmented counterpart (see~\figurename{~\ref{fig:motivation}} (b)$\to$(c)). Accordingly, we design the feature alignment loss to measure the difference between the two features.
We have evaluated our ProSFDA against other SFDA methods on a multi-domain joint optic disc (OD) and optic cup (OC) segmentation dataset and a multi-domain gray matter (GM) and white matter (WM) segmentation dataset.
To summarise, the contributions of this work are three-fold.
\begin{itemize}
\item[$\bullet$] We highlight the domain discrepancy issue existed under the SFDA setting and introduce PLS to address it from the perspective of estimating a domain-aware visual prompt via minimizing the statistic alignment loss.
\item[$\bullet$] We develop FAS to force the model to extract compact features from altered target-domain images and diminish the impact of instance variations.
\item[$\bullet$] Our ProSFDA achieves superior performance against other SFDA methods on two multi-domain medical image segmentation benchmarks.
\end{itemize}

\section{Related Work}
\subsection{Source-free Domain Adaptation}
SFDA methods can be roughly categorized into pseudo label-based approaches and prior-guided approaches.
With the recent advance of noisy label learning~\cite{song2022learning}, many \textbf{pseudo label-based} methods~\cite{chen_source-free_2021,xu_denoising_2022,dong_confident_2021,chu_denoised_2022,ding_source-free_2022,yang_exploiting_2021} have been developed to correct the pseudo labels generated by the source model, and thereby fine-tune the source model using the target-domain images with their corrected pseudo labels.
For example, Chen \etal~\cite{chen_source-free_2021} proposed a pixel- and class-level pseudo label 
denoising method, which corrects pseudo labels using the uncertainty maps and class prototypes produced by the source model.
Xu \etal~\cite{xu_denoising_2022} employed uncertainty maps to estimate the joint distribution matrix between the observed and latent labels for pseudo label refinement.
These methods work well when the estimated probability maps are reliable. However, when the domain discrepancy is large, the source model may produce overconfident wrong probability maps on target images, leading to substantial performance degradation on the target domain.
\textbf{Prior-guided methods}, such as AdaEnt~\cite{bateson_source-relaxed_2020} and AdaMI~\cite{bateson_source-free_2021}, restore the source-domain information using an additional prior knowledge prediction network.
They adopt the entropy loss and prior consistency loss to transfer the prior knowledge across domains.
Besides, based on the prior that the style information, which is the major cause of domain discrepancy, is mostly low frequency signals in images, researchers attempted to recover the style of source-domain images using the source model and target-domain images ~\cite{yang_source_2022,liu_source-free_2022,liu_source-free_2021,li_model_2020}. Yang \etal~\cite{yang_source_2022} proposed a Fourier style mining generator to restore source-like images based on the statistical information of the pre-trained source model and the mutual Fourier Transform.
Although they can achieve knowledge transfer in the absence of source-domain data, most of these methods heavily rely on the consistent prior assumption and the additional prior knowledge predictor, whose reliability still can be influenced by the domain discrepancy.

By contrast, the proposed ProSFDA employs a learnable domain-aware prompt to estimate the domain discrepancy using only the source model and target-domain images, and also aligns the features of altered target-domain images and their style-augmented counterparts, making the model robust to the instance variations of target images.

\subsection{Prompt Learning}
Prompt was initially designed to preappend as language instruction to the input text for fine-tuning large-scale Natural Language Processing (NLP) models to adapt to the downstream tasks~\cite{liu2021pre}.
Recently, with the success of prompt in NLP, its visual counterpart, visual prompt, has been increasingly studied~\cite{jia_visual_2022,bahng_visual_2022,zhang_amortized_2022,zheng_prompt_2022}.
Jia \etal~\cite{jia_visual_2022} introduced a small amount of trainable parameters in the input space while keeping the pre-trained backbone frozen, and thus achieved significant performance gains over other model tuning protocols.
Bahng \etal~\cite{bahng_visual_2022} proposed a visual prompt in the form of pixels to learn a task-specific image perturbation, and prompted a frozen model with this perturbation to perform a new task.
Due to its flexible form, the learned prompt can successfully bridge the gap between the frozen model and downstream data / tasks.

Our ProSFDA employs the prompt in the form of a learnable visual image and adds it to target images to resolve the domain discrepancy issue for source-free domain adaptive medical image segmentation.
In our solution, the prompt is optimized in an unsupervised manner via aligning the BN statistics in the source model and running source model (\ie, the model that takes altered target images as inputs).

\begin{figure*}[t]
  \centering
  \includegraphics[width=0.9\textwidth]{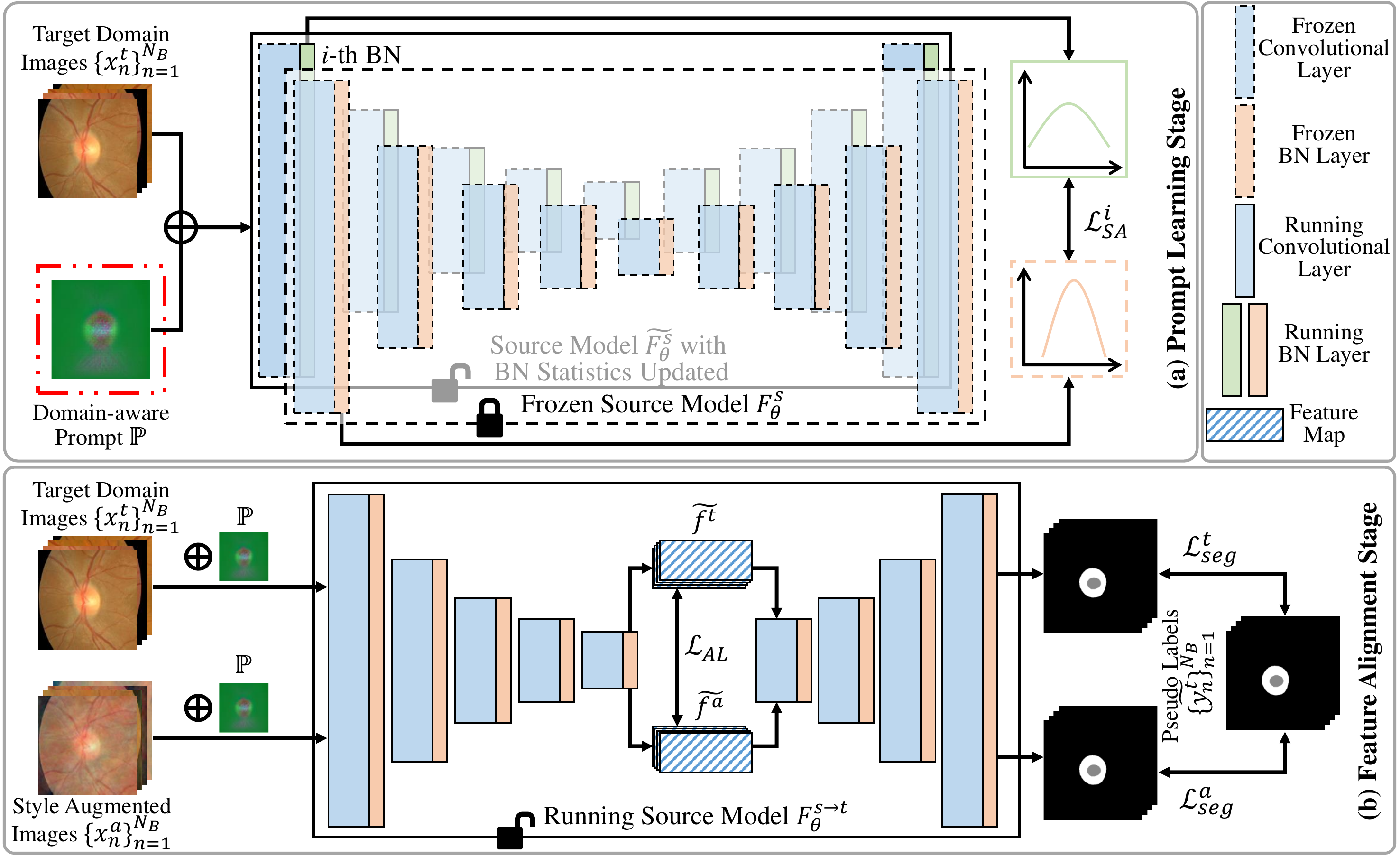}
  \caption{
Diagram of our ProSFDA. It is composed of (a) Prompt Learning Stage (PLS) and (b) Feature Alignment Stage (FAS). In PLS, a zero-initialized learnable prompt is added to each input target image, and adjusted by minimizing $\mathcal{L}_{SA}$. Note that the altered target images $\{\widetilde{x_n^t}\}_{n=1}^{N_B}$ are fed to the source model $\widetilde{F_{\theta}^s}$ that only BN statistics can be updated. In FAS, $\mathcal{L}_{seg}$ and $\mathcal{L}_{AL}$ are minimized to adjust the model toward extracting compact features. 
  }
  \label{fig:overview}
\end{figure*}

\section{Method}
\subsection{Problem Definition and Method Overview}
Given a segmentation model $F_{\theta}^s: x\to y$ pre-trained on the data from unseen source domain(s) $\mathbb{D}^s=\{(x^s_n, y^s_n)_{n=1}^{N^s}\}$, and a set of unlabeled data $\{x_n^t\}_{n=1}^{N^t}$ from target domain $\mathbb{D}^t$, where $N^s$ and $N^t$ represent the amount of the source images and target images, $x_n^t \in \mathbb{R}^{H\times W\times C}$ is the $n$-th target-domain image with $C$ channels and size of $H\times W$.
Our goal is to adapt the model $F_{\theta}^s: x\to y$ using $\{x_n^t\}_{n=1}^{N^t}$, and obtain the model $F_{\theta}^{s\to t}: x\to y$, which can perform well on the target domain $\mathbb{D}^t$.

Our ProSFDA can be applied to any pre-trained segmentation model with BN layers.
In this paper, we adopt a U-shape~\cite{falk_u-net_2019} segmentation model pre-trained on source data as the source model. 
The training of ProSFDA contains two sequentially conducted stages, \ie, Prompt Learning Stage (PLS) and Feature Alignment Stage (FAS).
During PLS, the zero-initialized domain aware prompt $\mathbb{P}$ is adjusted to estimate the domain discrepancy.
During FAS, $F_{\theta}^s$ is adjusted to align the altered target-domain image and its counterpart with a different style.
With the estimated $\mathbb{P}$ and the adjusted $F_{\theta}^{s\to t}$, we can perform prediction for target images from $\mathbb{D}^t$.
The diagram of our ProSFDA is shown in~\figurename{~\ref{fig:overview}}. We now delve into its details.

\subsection{Prompt Learning Stage}
\noindent
\textbf{Domain-aware Prompt Definition.}
The goal of domain-aware prompt $\mathbb{P}$ is to alter $x_n^t$ so that the altered target-domain image $\widetilde{x_n^t}$ can be treated as $x^s$ by $F_{\theta}^s$.
We design the learnable domain-aware prompt $\mathbb{P}\in \mathbb{R}^{H\times W\times C}$
so that $\mathbb{P}$ can be directly added to $x_n^t$.
The altered target image $\widetilde{x_n^t}$ can be represented as
\begin{equation}
\begin{aligned}
\widetilde{x_n^t} = x_n^t + \mathbb{P}
\end{aligned}
\end{equation}

\noindent
\textbf{Domain-aware Prompt Estimation.}
The altered target images batch $\{\widetilde{x_n^t}\}_{n=1}^{N_B}$, where $N_B$ represents the batch size, are fed to the running source model $\widetilde{F_{\theta}^s}$ with only BN statistics updated during forward propagation.
Due to the existence of domain discrepancy, the calculated BN statistics (\ie, $\widetilde{\mu}$ and $\widetilde{\sigma}$) are different from the BN statistics (\ie, $\mu$ and $\sigma$) stored in the frozen source model $F_{\theta}^s$.
We design an L1 norm based statistic alignment loss $\mathcal{L}_{SA}$ to measure the domain discrepancy, shown as follows.
\begin{equation}
\begin{aligned}
\mathcal{L}_{SA} = \sum\mathcal{L}_{SA}^i = \sum (\left | \mu_i - \widetilde{\mu_i} \right | + \alpha \times \left | \sigma_i - \widetilde{\sigma_i} \right |)
\end{aligned}
\end{equation}
where $i$ represents $i$-th BN layer in the model, and $\alpha$ is a balanced factor, which is set to $0.01$ in our experiments.

Note that $\mathbb{P}$ is zero-initialized so that the measured domain discrepancy can be reliable in the beginning.
During back propagation, only $\mathbb{P}$ is adjusted to minimize $\mathcal{L}_{SA}$, whereas the other learnable parameters in $\widetilde{F_{\theta}^s}$ are frozen.
After each iteration, $\widetilde{F_{\theta}^s}$ is initialized as $F_{\theta}^s$ again, so that the BN statistics will not momentum updated to fit $\{\widetilde{x_n^t}\}_{n=1}^{N_B}$.

\subsection{Feature Alignment Stage}
\noindent
\textbf{Frequency-based Style Augmentation.}
We adopt low-frequency replacement within a batch of target-domain images to increase the style diversity of them.
Given a target image batch $\{x_n^t\}_{n=1}^{N_B}$, first we permute it to match $x_n^t$ with $x_p^t$, which provides a different reference image style.
Then we transform the image $x_n^t$ and its permuted counterpart $x_p^t$ to the frequency domain using Fast Fourier Transform (FFT).
\begin{equation}
\begin{aligned}
\{Amp(x_n^t), Pha(x_n^t)\} = FFT(x_n^t) 
\end{aligned}
\end{equation}
where $Amp(x_n^t)$ and $Pha(x_n^t)$ represent the amplitude map and phase map of $x_n^t$ respectively.

The low frequency component of $x_n^t$ can be represented as $Low(Amp(x_n^t);\beta)$, where $\beta$ is the cut-off ratio between low and high frequency components, and is randomly selected from $(0,0.15]$.
Based on the assumption that the style information is embedded in low frequency components~\cite{hu_domain_2022,yang_fda_2020,huang2021rda}, we replace $Low(Amp(x_n^t);\beta)$ with $Low(Amp(x_p^t);\beta)$ to alter the style of $x_n^t$.
After that, we conduct Inverse Fast Fourier Transform (IFFT) to recover the augmented image $x_n^a$ with a different style.

\noindent
\textbf{Feature Alignment.}
Using the target image $x_n^t$ and its counterpart $x_n^a$ with a different style, it is expected the extracted features from these paired images are the same, since only their styles are different, which should be depressed for the segmentation task.
We therefore design an L1 norm based feature alignment loss to measure the mismatch of the extracted features from the paired images, shown as follows.
\begin{equation}
\begin{aligned}
\mathcal{L}_{AL} = \left | \widetilde{f^t} - \widetilde{f^a} \right |
\end{aligned}
\end{equation}
where $\widetilde{f^t}$ and $\widetilde{f^a}$ represent the features extracted from $\widetilde{x^t}$ and $\widetilde{x^a}$ by the encoder of the segmentation model respectively.

If we adjust the model to only minimize $\mathcal{L}_{AL}$, the model may collapse since there is no constraint to ensure $\widetilde{f^t}$ and $\widetilde{f^a}$ are fit for the segmentation task.
Therefore, we use the pseudo label $\widetilde{y^t}$ of $\widetilde{x^t}$ generated by $F_{\theta}^s$ as additional supervision to avoid it.
We adopt the cross-entropy loss as the segmentation loss
\begin{equation}
\begin{aligned}
\mathcal{L}_{seg} = \mathcal{L}_{seg}^t + \mathcal{L}_{seg}^a = 
\mathcal{L}_{ce}(\widetilde{y^t}, \hat{y^t}) +\mathcal{L}_{ce}(\widetilde{y^t}, \hat{y^a})
\end{aligned}
\end{equation}
where $\hat{y^t}$ and $\hat{y^a}$ represent the prediction for $\widetilde{x^t}$ and $\widetilde{x^a}$ respectively, and
\begin{equation}
\begin{aligned}
\mathcal{L}_{ce}(\widetilde{y}, \hat{y}) = -(\widetilde{y} \log \hat{y} +\left(1-\widetilde{y}\right) \log \left(1-\hat{y}\right))
\end{aligned}
\end{equation}

The total loss function $\mathcal{L}_{FAS}$ during the fine-tuning process can be represented as
\begin{equation}
\begin{aligned}
\mathcal{L}_{FAS} = \mathcal{L}_{seg} + \gamma \times \mathcal{L}_{AL}
\end{aligned}
\end{equation}
where $\gamma$ is a weight factor, which is set to 0.1 in our experiments.

\subsection{Training and Test}
\noindent \textbf{Training.} During training, the domain-aware prompt $\mathbb{P}$ is estimated in PLS by minimizing $\mathcal{L}_{SA}$.
Using $\mathbb{P}$, the target-domain images can be altered as $\widetilde{x^t}$.
Then the pseudo label $\widetilde{y^t}$ for $\widetilde{x^t}$ can be generated using the pre-trained source model $F_{\theta}^s$.
In FAS, the source model $F_{\theta}^s$ is adjusted to minimize $\mathcal{L}_{seg}$ and $\mathcal{L}_{AL}$ that measure the mismatch between $\widetilde{x^t}$ and its counterpart $\widetilde{x^a}$ with a different style.
Finally, we can get the adapted model $F_{\theta}^{s\to t}$.

\noindent \textbf{Test.} During test, given a test image $x^t$, we alter it with $\mathbb{P}$ as $\widetilde{x^t}$. Then we feed $\widetilde{x^t}$ to $F_{\theta}^{s\to t}$ to generate the final segmentation result $\hat{y^t}$.

\section{Experiments}
\subsection{Datasets}
\noindent \textbf{The RIGA+ dataset}~\cite{hu_domain_2022,almazroa2018retinal,decenciere2014feedback}
is a multi-domain joint OC/OD segmentation dataset annotated by six ophthalmologists.
To reduce the annotator bias~\cite{liao2021modeling} among datasets from different domains, only rater 1’s annotations are used to train and evaluate all the algorithms.
Totally five domains are included in this dataset, \ie, BinRushed, Magrabia, BASE1, BASE2, and BASE3.
There are 195, 95, 173, 148, and 133 labeled data from these domains.
Of the latter three domains, there are additional 227, 238, and 252 unlabeled fundus images.
We chose the first two domains, \ie, BinRushed and Magrabia, as source domains, and the latter three domains, \ie, BASE1, BASE2, and BASE3, as target domains respectively.

\noindent \textbf{The SCGM dataset}~\cite{prados2017spinal}
is another multi-domain dataset annotated by four expert raters for joint GM/WM segmentation.
We only use rater 1’s annotations to train and evaluate all the algorithms.
There are four domains included in this dataset, \ie, Site 1 (UCL), Site 2 (Montreal), Site 3 (Zurich), and Site 4 (Vanderbilt).
Each domain contains 10 labeled 3D MR volumes and 10 unlabeled 3D MR volumes.
We chose Site 1 as the target domain and the left three domains as source domains.

\subsection{Experimental Setup}
\noindent \textbf{Implementation Details.}
The images in each segmentation task were normalized by subtracting the mean and dividing by the standard deviation.
For all experiments, the mini-batch size was set to 16.
The input image size was set to $512\times 512$ for joint OC/OD segmentation, and $128\times 128$ for joint GM/WM segmentation.
The images in the RIGA+ dataset were center-cropped to $800\times 800$ before resizing to $512\times 512$, whereas in the SCGM dataset, the images were normalized to a uniform in-plane pixel size of $0.25mm \times 0.25mm$ and center-cropped to $128\times 128$.
The SGD algorithm with a momentum of 0.99 was adopted as the optimizer.
The initial learning rate was set to $lr_0=0.01$ in PLS, and $lr_0=0.001$ in FAS.
It was decayed according to the polynomial rule $lr=lr_0\times(1-t/T)^{0.9}$, where $t$ is the current epoch and $T=100$ is the maximum epoch for both PLS and FAS.
The source model was trained using labeled source data in advance, and kept as the same for all experiments.
All SFDA methods were trained using the unlabeled target data and evaluated on the labeled data.
All experiments were implemented using the PyTorch~\cite{paszke2019pytorch} framework and performed with one NVIDIA 2080Ti GPU.

\noindent \textbf{Evaluation Metrics.}
Dice Similarity Coefficient ($\mathcal{D}$, \%), which characterizes the accuracy of the predicted masks, was adopted as the evaluation metric to measure the segmentation performance. 
Higher $\mathcal{D}$ represents better segmentation performance.

\begin{table*}[]
\setlength\tabcolsep{10pt}
\centering
\caption{
Average performance (mean $\pm$ standard deviation) of three trials of our ProSFDA, ten competing methods, and three variants of ProSFDA in joint OC/OD segmentation. The best results among three baseline methods are highlighted with \underline{underline}. The best results among six SFDA methods are highlighted with \textbf{bold}.
}
\label{tab:RIGA+-compare}
\begin{tabular}{l|c|c|c|c|c|c}
\hline \hline
\multirow{2}{*}{Methods} & \multicolumn{2}{c|}{BASE1} & \multicolumn{2}{c|}{BASE2} & \multicolumn{2}{c}{BASE3} \\ \cline{2-7} 
                  & $\mathcal{D}_{OD}$ & $\mathcal{D}_{OC}$ & $\mathcal{D}_{OD}$ & $\mathcal{D}_{OC}$ & $\mathcal{D}_{OD}$ & $\mathcal{D}_{OC}$ \\
\hline
Intra-Domain & 94.71$\pm$0.07 & 84.07$\pm$0.35 & \underline{94.84$\pm$0.18} & \underline{86.32$\pm$0.14} & \underline{95.40$\pm$0.05} & \underline{87.34$\pm$0.11} \\ \hline
w/o DA & \underline{95.31$\pm$0.10} & \underline{84.42$\pm$0.09} & 92.37$\pm$0.63 & 78.67$\pm$1.67 & 94.10$\pm$0.41 & 83.12$\pm$1.03 \\ \hline
Self-Training & 95.16$\pm$0.14 & 83.83$\pm$0.08 & 91.22$\pm$0.10 & 75.42$\pm$0.20 & 94.11$\pm$0.12 & 82.50$\pm$0.07 \\ \hline
\hline
BEAL~\cite{wang2019boundary} & 95.80$\pm$0.07 & 86.00$\pm$0.18 & 94.53$\pm$0.31 & 83.12$\pm$0.30 & 95.49$\pm$0.10 & 85.50$\pm$0.79 \\ \hline
DoCR~\cite{hu_domain_2022} & 96.15$\pm$0.11 & 86.03$\pm$0.07 & 95.90$\pm$0.13 & 85.68$\pm$0.39 & 96.10$\pm$0.02 & 86.58$\pm$0.21 \\ \hline
\hline
AdaEnt~\cite{bateson_source-relaxed_2020} & 94.46$\pm$0.03 & 82.71$\pm$0.06 & 92.77$\pm$0.02 & 77.79$\pm$0.03 & 93.72$\pm$0.03 & 81.87$\pm$0.04 \\ \hline
AdaMI~\cite{bateson_source-free_2021} & 94.50$\pm$0.06 & 82.80$\pm$0.19 & 92.72$\pm$0.02 & 78.86$\pm$0.19 & 93.65$\pm$0.06 & 82.71$\pm$0.11 \\ \hline
DPL~\cite{chen_source-free_2021} & 95.18$\pm$0.04 & 83.90$\pm$0.20 & 92.91$\pm$0.20 & 79.06$\pm$0.42 & 94.15$\pm$0.11 & 82.62$\pm$0.08 \\ \hline
U-D4R~\cite{xu_denoising_2022} & 95.17$\pm$0.07 & 83.94$\pm$0.12 & 93.57$\pm$0.02 & 78.79$\pm$0.02 & 94.27$\pm$0.02 & 83.30$\pm$0.03 \\ \hline
FSM~\cite{yang_source_2022} & 94.96$\pm$0.63 & 84.30$\pm$1.47 & 93.10$\pm$0.32 & 81.39$\pm$0.91 & 94.41$\pm$0.98 & 83.21$\pm$1.92 \\ \hline
Ours & \textbf{95.29$\pm$0.12} & \textbf{85.61$\pm$0.24} & \textbf{94.71$\pm$0.01} & \textbf{85.33$\pm$0.08} & \textbf{95.47$\pm$0.01} & \textbf{85.53$\pm$0.15} \\ \hline
\hline
Ours w/o FAS & 95.43$\pm$0.00 & 85.02$\pm$0.00 & 94.89$\pm$0.00 & 83.97$\pm$0.00 & 95.21$\pm$0.00 & 85.07$\pm$0.00 \\ \hline
Ours w/o PLS & 95.48$\pm$0.03 & 85.33$\pm$0.12 & 94.73$\pm$0.01 & 83.36$\pm$0.30 & 95.36$\pm$0.02 & 85.15$\pm$0.12 \\ \hline
Ours w/o PLS* & 95.23$\pm$0.02 & 84.67$\pm$0.27 & 92.31$\pm$0.13 & 78.15$\pm$0.36 & 94.64$\pm$0.04 & 83.03$\pm$0.18 \\ \hline
\hline
\end{tabular}
\end{table*}

\subsection{Experimental Results}
We compared the proposed ProSFDA with 
the `Intra-domain’ setting (\ie, training and testing on the labeled data from the target domain using 3-fold cross-validation), 
`w/o DA' baseline (\ie, training on the data aggregated from source domains and testing directly on the target domain),
`Self-Training' baseline (\ie, training on the target-domain data labeled by the pre-trained source model and testing on the target domain),
and five SFDA methods, including 
two pseudo label-based methods, \ie, DPL~\cite{chen_source-free_2021} and U-D4R~\cite{xu_denoising_2022}, and three prior-guided methods, \ie, AdaEnt~\cite{bateson_source-relaxed_2020}, AdaMI~\cite{bateson_source-free_2021}, and FSM~\cite{yang_source_2022}.
The performance of two UDA methods, \ie, BEAL~\cite{wang2019boundary}, and DoCR~\cite{hu_domain_2022}, are also displayed for reference.
We re-implemented all the competing methods using the same backbone as our ProSFDA, and reported their best performance.

\noindent
\textbf{Comparison on RIGA+ Dataset.}
The Dice Similarity Coefficient obtained by our model and other competing methods were listed in~\tablename{~\ref{tab:RIGA+-compare}}.
The domain discrepancy can be reflected by the performance gap between `Intra-Domain' and `w/o DA'.
It reveals that the domain discrepancy between source domains and `BASE1' is relatively smaller, whereas the domain discrepancies between source domains and either `BASE2' or `BASE3' are relatively larger, especially in `BASE2'.
It shows that AdaEnt and AdaMI failed in the joint OC/OD segmentation task, and their performances are even inferior to `w/o DA'.
We assume this can be attributed to that the class ratio of OC/OD can be largely different among different domains.
Even though, our ProSFDA significantly outperforms all SFDA methods, and is even comparable to BEAL, a UDA method that can access source-domain data during training.
It demonstrates the effectiveness of the proposed ProSFDA.

\begin{figure*}[t]
  \centering
  \includegraphics[width=1.0\textwidth]{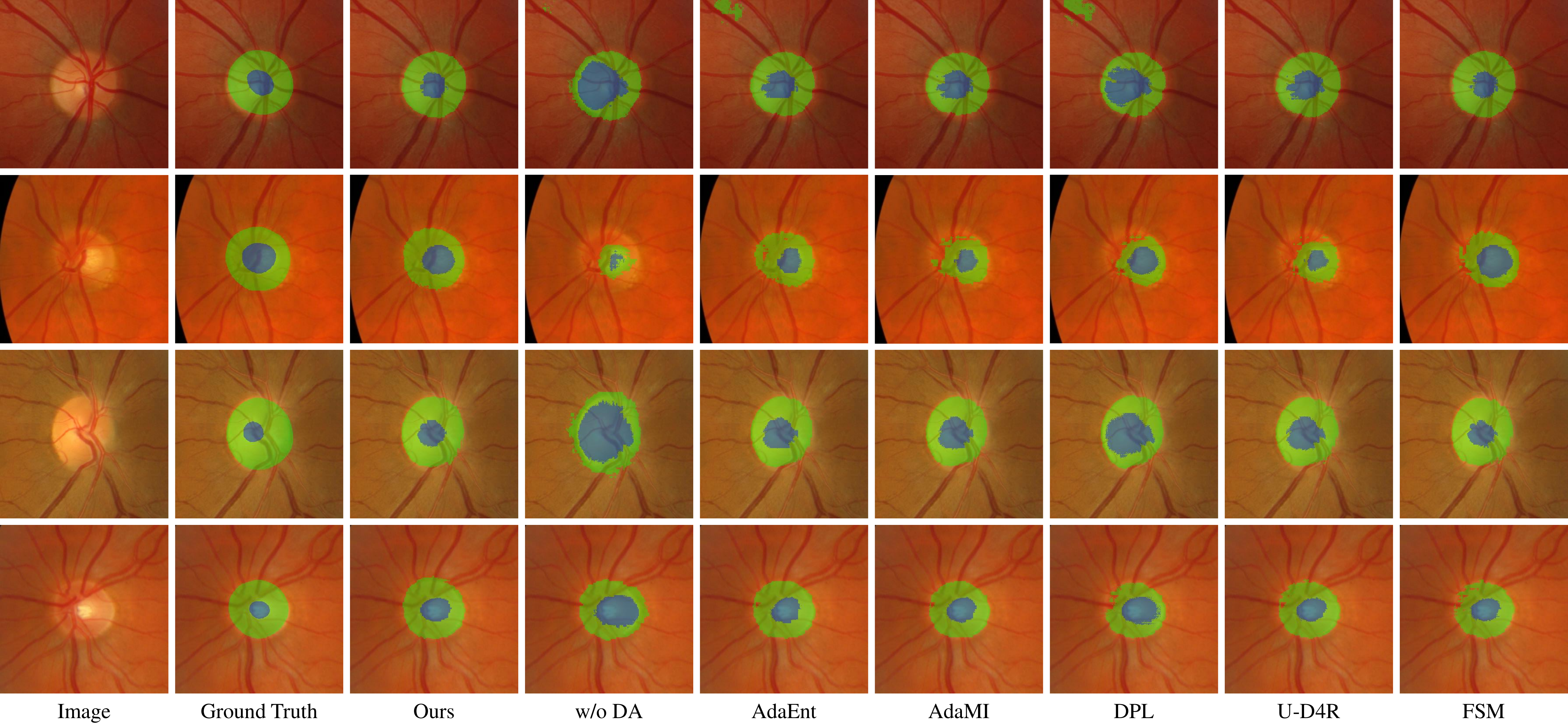}
  \caption{
Visualization of segmentation masks predicted by our ProSFDA and six competing methods, together with ground truth.
  }
  \label{fig:compare-vis}
\end{figure*}

We visualize the segmentation maps predicted by our ProSFDA and other five SFDA methods in~\figurename{~\ref{fig:compare-vis}}.
Also, the ground truth segmentation maps and the results of `w/o DA' are also displayed for reference.
It shows that our ProSFDA can produce the most similar segmentation maps compared to the ground truth, confirming the superiority of our method.

\begin{table}[]
\setlength\tabcolsep{13pt}
\centering
\caption{
Average performance (mean $\pm$ standard deviation) of three trials of our ProSFDA, ten competing methods in spin core gray matter segmentation. The best results among three baseline methods are highlighted with \underline{underline}. The best results among six SFDA methods are highlighted with \textbf{bold}.
}
\label{tab:SCGM-compare}
\begin{tabular}{l|c|c}
\hline \hline
\multirow{2}{*}{Methods} & \multicolumn{2}{c}{Target Domain} \\ \cline{2-3} 
                  & $\mathcal{D}_{GM}$ & $\mathcal{D}_{WM}$ \\
\hline
Intra-Domain & 93.35$\pm$0.24 & 44.02$\pm$2.52 \\ \hline
w/o DA & 95.40$\pm$0.06 & 64.96$\pm$0.62 \\ \hline
Self-Training & \underline{95.46$\pm$0.09} & \underline{65.15$\pm$0.27} \\ \hline
\hline
BEAL~\cite{wang2019boundary} & 96.32$\pm$0.07 & 72.82$\pm$0.49 \\ \hline
DoCR~\cite{hu_domain_2022} & 96.82$\pm$0.04 & 75.11$\pm$0.34 \\ \hline
\hline
AdaEnt~\cite{bateson_source-relaxed_2020} & 96.10$\pm$0.09 & 68.56$\pm$0.89 \\ \hline
AdaMI~\cite{bateson_source-free_2021} & 96.16$\pm$0.09 & 69.29$\pm$0.51 \\ \hline
DPL~\cite{chen_source-free_2021} & 95.77$\pm$0.06 & 67.73$\pm$0.49 \\ \hline
U-D4R~\cite{xu_denoising_2022} & 95.89$\pm$0.04 & 68.12$\pm$0.53 \\ \hline
FSM~\cite{yang_source_2022} & 96.32$\pm$0.06 & 70.67$\pm$0.43 \\ \hline
Ours & \textbf{96.42$\pm$0.05} & \textbf{72.76$\pm$0.45} \\ \hline
\hline
\end{tabular}
\end{table}

\noindent
\textbf{Comparison on SCGM Dataset.}
We also compare our ProSFDA with other methods on the SCGM dataset, as shown in~\tablename{~\ref{tab:SCGM-compare}}.
Since there are only a few slices in the target domain, the `Intra-Domain' performance is relatively low.
It shows the performance of `w/o DA' is much higher than `Intra-Domain', since there are more available labeled training data from source domains for `w/o DA'.
Even though, the UDA methods can achieve better performance than `w/o DA', demonstrating that diminishing the domain discrepancy can improve the performance on the target domain.
Also, it shows that the SFDA methods can improve the performance compared to `w/o DA' and `Self-Training'.
Nonetheless, our ProSFDA beats all the competing SFDA methods.
This observation is consistent with what we observed in~\tablename{~\ref{tab:RIGA+-compare}}.

\subsection{Ablation Analysis}
Both PLS and FAS play an essential role in the proposed ProSFDA,
estimating the domain discrepancy and enhancing the feature compactness independently.
We conducted ablation studies on the RIGA+ dataset to investigate the effectiveness of them respectively.
And the joint OC/OD segmentation task on the target domain `BASE2' is chosen as a case study to analyze PLS and FAS independently.

\noindent
\textbf{Contributions of PLS and FAS.}
To evaluate the contributions of PLS and FAS, we compared the performance of ProSFDA with its variants that use only one module, \ie, `Ours w/o FAS' and `Ours w/o PLS'.
`Ours w/o FAS' denotes the variant that only estimates the domain-aware prompt using $\mathcal{L}_{SA}$. During inference, the test image is altered with the estimated prompt and then fed to the source model $F_{\theta}^s$ to produce the segmentation result for evaluation.
In `Ours w/o PLS', the estimated prompt is used to generate the pseudo labels as the supervision during FAS, whereas the input target-domain images and their counterparts are not altered with prompt.
We also compared to a variant, represented by `Ours w/o PLS*', that uses the pseudo labels generated by `w/o DA' as supervision during FAS.
The performance of our ProSFDA and its three variants are given in~\tablename{~\ref{tab:RIGA+-compare}}.
It shows that, compared to `Ours w/o PLS', when not using prompt to generate pseudo labels as supervision in FAS, the performance of `Ours w/o PLS*' drops substantially, but is still better than `Self-Training'.
It reveals that the estimated domain-aware prompt is beneficial to the segmentation performance.
On the other hand, compared to `Ours', when FAS is removed, the performance of `Ours w/o FAS' drops significantly.
It demonstrates that feature alignment can also benefit the segmentation performance.
Also, it shows the standard deviation of three trials of `Ours w/o FAS' is zero.
We believe this is due to only a few learnable parameters (\ie, prompt) can be adjusted in PLS, and the initialization of three trials is the same source model.
Moreover, it reveals that our ProSFDA outperforms all of its variants.
The results confirm that both PLS and FAS contribute to the final results.

\begin{table}[]
\setlength\tabcolsep{13pt}
\centering
\caption{
Average performance (mean $\pm$ standard deviation) of three trials of ProSFDA with only PLS when adopting different operations on `BASE2' of joint OC/OD segmentation. The `Frequency' represents the operations conducted in the frequency domain. The best results are highlighted with \textbf{bold}.
}
\label{tab:PLS-operation}
\begin{tabular}{c|c|c}
\hline \hline
{\multirow{2}{*}{Operation}} & \multicolumn{2}{c}{Average} \\ \cline{2-3} 
                           & $\mathcal{D}_{OD}$ & $\mathcal{D}_{OC}$  \\ \hline
Frequency $\times$ & 83.51$\pm$0.55 & 69.81$\pm$0.15 \\ \hline
Frequency + & 94.53$\pm$0.00 & 82.29$\pm$0.00 \\ \hline
$\times$ & 91.81$\pm$0.00 & 79.53$\pm$0.00 \\ \hline
+ & \textbf{94.89$\pm$0.00} & \textbf{83.97$\pm$0.00} \\ \hline
\hline
\end{tabular}
\end{table}

\begin{figure}[]
  \centering
  \includegraphics[width=0.48\textwidth]{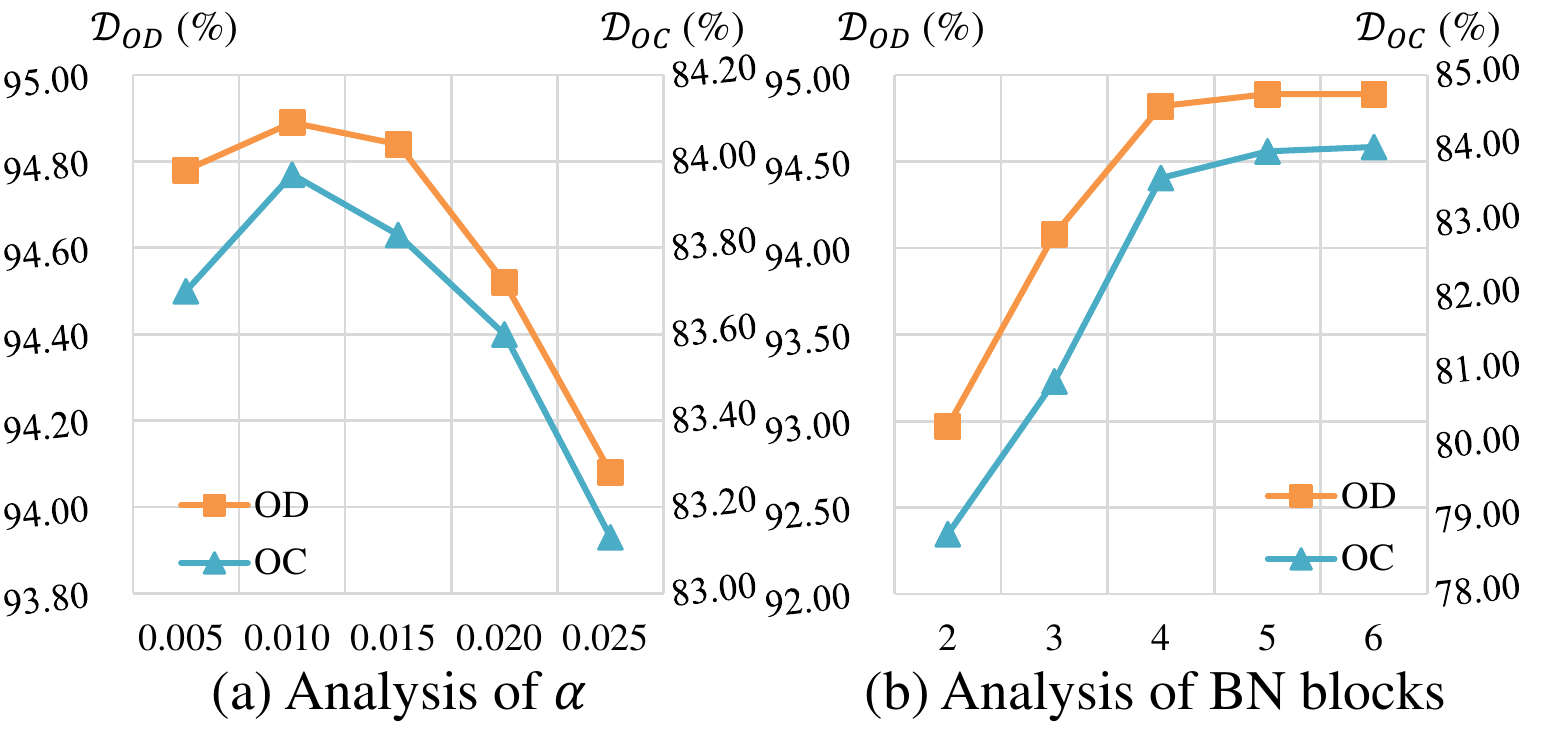}
  \caption{
Performance of ProSFDA with only PLS on `BASE2' of joint OC/OD segmentation when setting (a) $\alpha$ and (b) BN blocks to different values.
  }
  \label{fig:prompt-hyper}
\end{figure}

\begin{figure}[]
  \centering
  \includegraphics[scale=0.2]{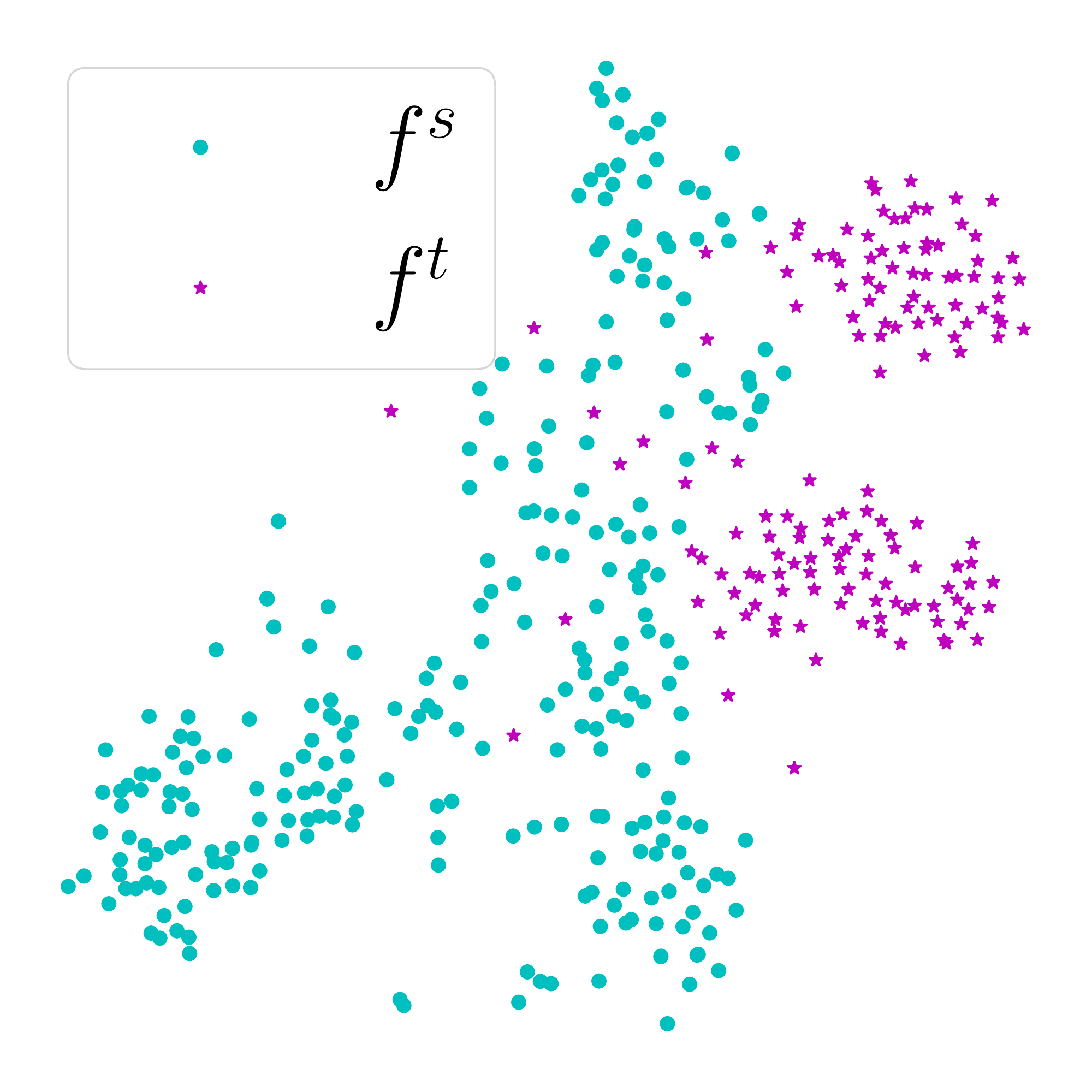}
  \includegraphics[scale=0.2]{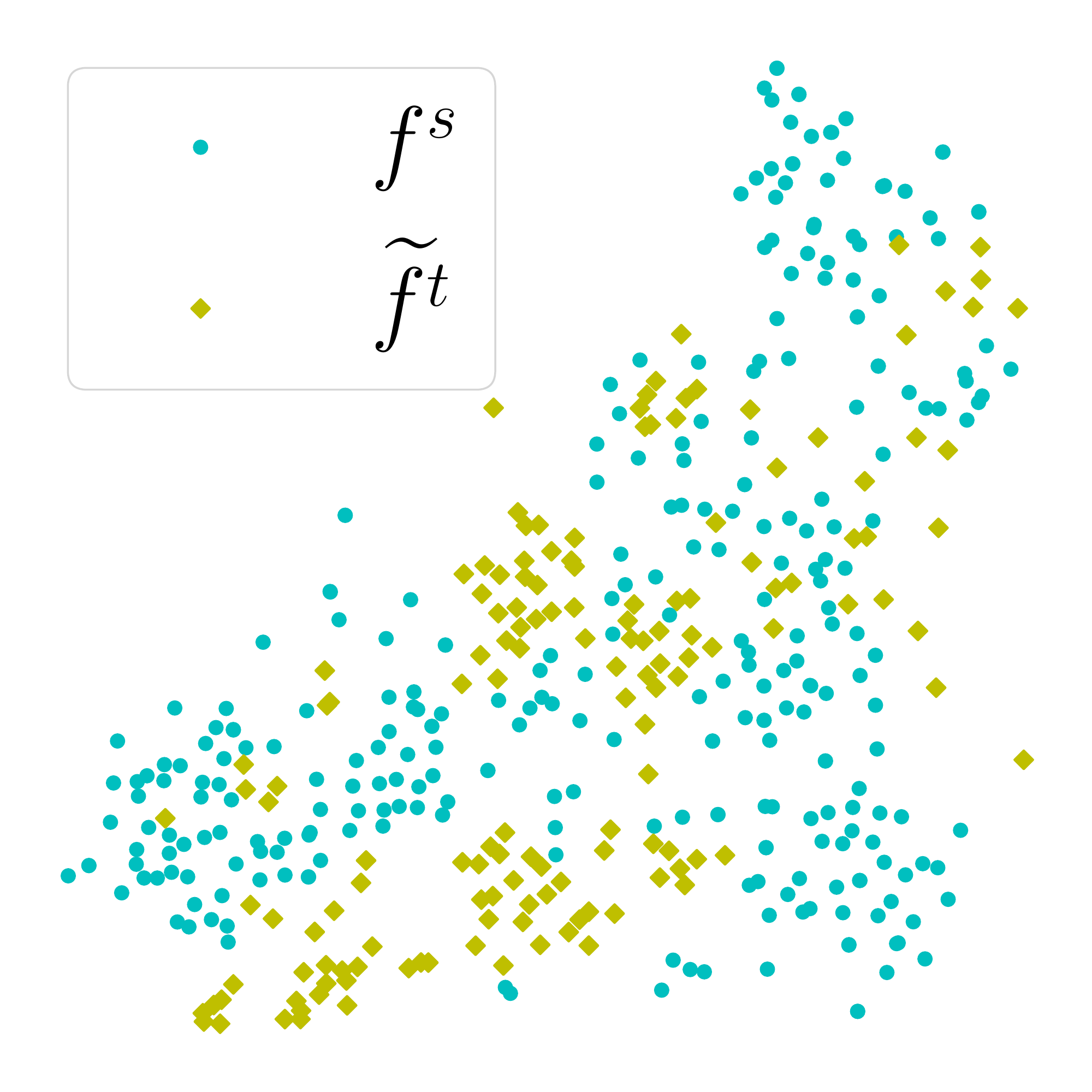}
  \caption{
t-SNE visualization of features extracted from images in `BASE2' of joint OC/OD segmentation before ($f^t$) and after ($\widetilde{f^t}$) added with prompt $\mathbb{P}$. The features $f^s$ extracted from source images are also displayed for reference.
}
  \label{fig:prompt-tsne}
\end{figure}

\noindent
\textbf{Analysis of PLS.}
We investigate the effectiveness of different operations between the target image and prompt.
The performance on `BASE2' are reported in~\tablename{~\ref{tab:PLS-operation}}.
We also report the performance of different operations in the frequency domain, where the domain-sensitive and -insensitive information can be separate~\cite{yang_fda_2020,hu_domain_2022,huang2021rda}.
Note that for the operation `+', the prompt is zero-initialized, whereas for the operation `$\times$', the prompt is ones-initialized.
It shows that conducting `+' in the frequency domain is inferior to the spatial domain, and conducting `$\times$' in the frequency domain even gets worse performance.
We assume that the large dynamic range in the frequency domain makes it hard to estimate a proper prompt.
Whereas the dynamic range in the spatial domain is relatively small, making it easier to estimate the prompt.
It also reveals that using `+' is better than `$\times$'.
Therefore, we chose to directly add the prompt to the data.

We also analyze how many BN layers should be included in $\mathcal{L}_{SA}$. 
Since there are lots of BN layers (\ie, totally 43 BN layers) in the adopted segmentation network, we analyze the effect of them at the encoder/decoder block level, each of which contains $1\sim5$ BN layers, depending on the specific design.
The performance of using different amounts of BN blocks is shown in~\figurename{~\ref{fig:prompt-hyper}} (b).
It suggests that when up to 5 BN blocks (\ie, $i=36$ in our design), the performance no longer increases.
The performance against different choices of $\alpha$ in $\mathcal{L}_{SA}$ is shown in~\figurename{~\ref{fig:prompt-hyper}} (a).
It shows that $\alpha=0.01$ is the best choice.

To further investigate the domain-aware prompt, we visualize the distribution of the features extracted from the target images and their counterparts altered with the prompt using t-SNE in~\figurename{~\ref{fig:prompt-tsne}}.
It shows that the altered target images become indistinguishable from the source-domain images.
It demonstrates that using the estimated prompt to alter the target images can reduce the domain discrepancy, confirming the effectiveness of the proposed PLS.

\begin{table}[]
\centering
\caption{
Average performance (mean $\pm$ standard deviation) of three trials of ProSFDA and its variants on `BASE2' of joint OC/OD segmentation. ProSFDA with only PLS is adopted as `Baseline'. The best results are highlighted with \textbf{bold}.
}
\label{tab:FAS-loss}
\begin{tabular}{c|c|c|c|c}
\hline \hline
\multirow{2}{*}{Baseline} & \multicolumn{2}{c|}{$\mathcal{L}_{FAS}$} & \multicolumn{2}{c}{Average} \\ \cline{2-5}
           & $\mathcal{L}_{seg}$ & $\mathcal{L}_{AL}$ & $\mathcal{D}_{OD}$  & $\mathcal{D}_{OC}$  \\ \hline
\checkmark &            &          & 94.89$\pm$0.00 & 83.97$\pm$0.00  \\ \hline
\checkmark & \checkmark &          & \textbf{95.01$\pm$0.01} & 84.07$\pm$0.11 \\ \hline
\checkmark &            &\checkmark& 85.62$\pm$0.61 & 72.45$\pm$1.06 \\ \hline
\checkmark & \checkmark &\checkmark& 94.71$\pm$0.01 & \textbf{85.33$\pm$0.08} \\ \hline
\hline
\end{tabular}
\end{table}

\begin{figure}[]
  \centering
  \includegraphics[scale=0.2]{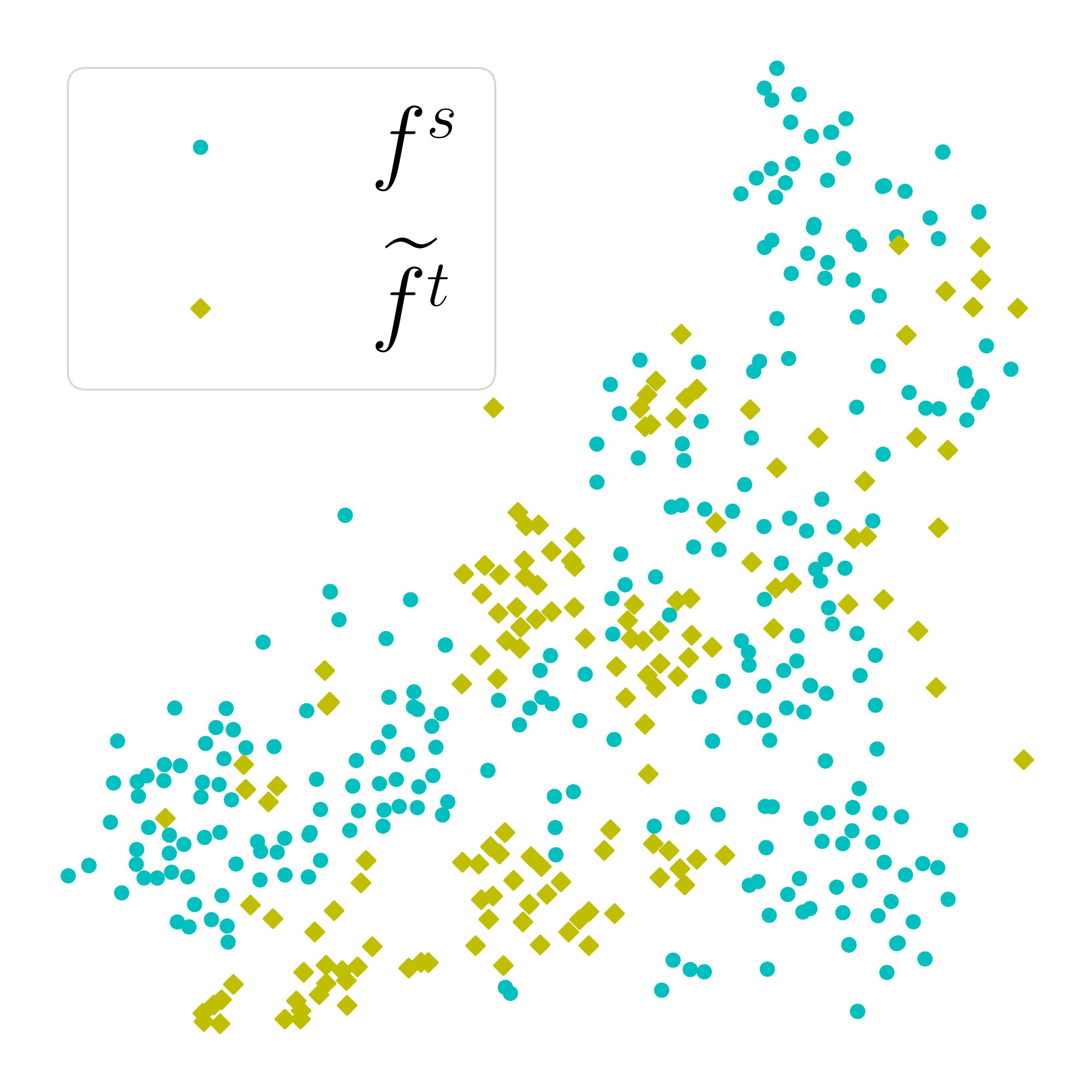}
  \includegraphics[scale=0.2]{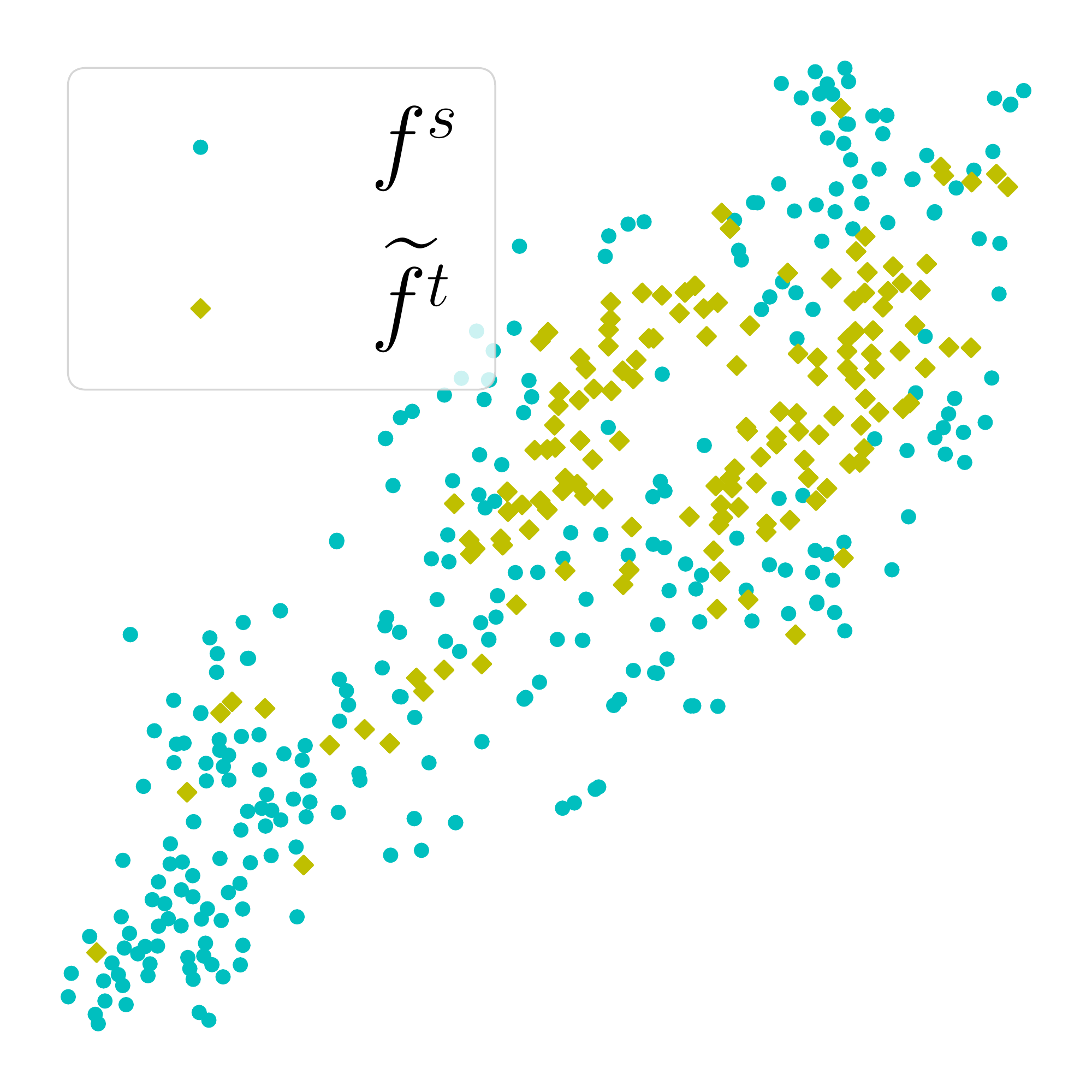}
  \caption{
t-SNE visualization of features extracted $\widetilde{f^t}$ from altered target-domain images in `BASE2' of joint OC/OD segmentation by $F_{\theta}^s$ (left) and $F_{\theta}^{s\to t}$ (right).
The features $f^s$ extracted from source images are also displayed for reference.
}
  \label{fig:FAS-tsne}
\end{figure}

\noindent
\textbf{Analysis of FAS.}
We analyze the impact of $\mathcal{L}_{seg}$ and $\mathcal{L}_{AL}$ in $\mathcal{L}_{FAS}$, as shown in~\tablename{~\ref{tab:FAS-loss}}.
The `Baseline' represent the performance of $F_{\theta}^s$ on $\widetilde{x^t}$.
It shows that self-training using $\widetilde{y^t}$ benefits the segmentation on the target domain, thanks to the diminished domain discrepancy in PLS.
However, if the model is updated to only minimize $\mathcal{L}_{AL}$, its performance drops significantly.
We assume that the model collapse to a trivial solution that the extracted features are domain-insensitive but not fit for the segmentation task.
It reveals that the overall performance of the combination of $\mathcal{L}_{seg}$ and $\mathcal{L}_{AL}$ is the best, confirming the effectiveness of the design of $\mathcal{L}_{FAS}$.

We visualize the distribution of the features of $\widetilde{x^t}$ extracted by the encoder of $F_{\theta}^s$ and $F_{\theta}^{s\to t}$ using t-SNE respectively, as shown in~\figurename{~\ref{fig:FAS-tsne}}.
It reveals that the features extracted by $F_{\theta}^{s\to t}$ are more compact than $F_{\theta}^s$, demonstrating that FAS can improve the compactness of the extracted features.

\section{Conclusion}
In this paper, we highlight the issue of domain discrepancy in SFDA, and proposed ProSFDA, a prompt learning based SFDA method.
Under ProSFDA, we introduce a learnable visual prompt to estimate the domain discrepancy and use the learned prompt to alter target domain images so that the altered target domain images can be treated as source domain ones by the frozen source model.
We also conduct feature alignment between the altered target domain images and their counterparts with different styles to adjust the model to extract compact features and diminish the impact of instance variation.
Our experimental results on two medical image segmentation benchmarks with multiple domains demonstrate that the proposed ProSFDA is superior to other SFDA methods, and even comparable with UDA methods.
However, since the estimation of the domain-aware prompt relies on the BN statistics alignment, our ProSFDA can only be applied to the networks equipped with BN layers currently. In our future work, we will investigate the domain-aware prompt estimation method that can be applied to other normalization techniques.

%% file: supplementary.tex
\section{Overview}
In this document, we provide the details of the source model we used in our experiment (Section~\ref{sec:SourceModel}), and complement more experimental results to supplement the main submission (Section~\ref{sec:Exp}).

\section{Details of Source Model}
\label{sec:SourceModel}
\subsection{Structure}
The source model is a U-shape architecture.
Its encoder is a tailored ResNet-34 whose average pooling layer and fully connected layer are replaced with a ReLU layer.
The tailored ResNet-34 contains a convolutional block and four residual blocks.
Symmetrically, the decoder is also composed of five convolutional blocks.
Skip connections are made from the each encoder block to the corresponding block in the decoder.

\subsection{Training}
For the joint OC/OD segmentation, the source model was trained using data from BinRushed and Magrabia.
For the joint GM/WM segmentation, the source model was trained using data from Site 2, Site 3, and Site 4.
The data preprocessing techniques (including cropping, and normalization), mini-batch size, and input patch size of training the source model are the same as training ProSFDA.
Several data augmentation techniques, including random cropping, rotation, scaling, flipping, adding Gaussian noise, and elastic deformation, were used to expand the training set.
The SGD algorithm with a momentum of 0.99 was adopted as the optimizer. 
The initial learning rate was set to 0.01 and decayed following the same polynomial rule as ProSFDA.
The maximum epoch was set to 100 for each segmentation task.

\section{Analysis of ProSFDA}
\label{sec:Exp}

\subsection{Analysis of Hyper-parameter in FAS}
\begin{figure}[]
  \centering
  \includegraphics[width=0.3\textwidth]{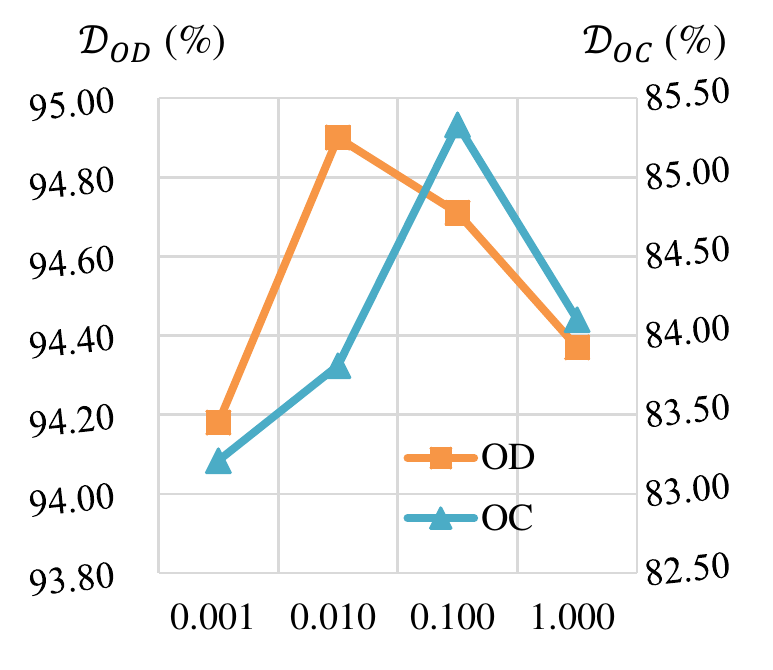}
  \caption{
Performance of ProSFDA on `BASE2' of joint OC/OD segmentation when setting $\gamma$ in $\mathcal{L}_{FAS}$ to different values.
  }
  \label{fig:gamma}
\end{figure}

To balance $\mathcal{L}_{seg}$ and $\mathcal{L}_{AL}$ in $\mathcal{L}_{FAS}$, we add a weight factor $\gamma$. We analyze the performance on `BASE2' against different selections of $\gamma$, as shown in~\figurename{~\ref{fig:gamma}}.
It shows that when $\gamma$ is set to 0.1, the best overall segmentation performance can be achieved.
We therefore set $\gamma$ to 0.1 in our experiments.

\subsection{Visualization of Domain-aware Prompt}
\begin{figure*}[]
  \centering
  \includegraphics[width=1\textwidth]{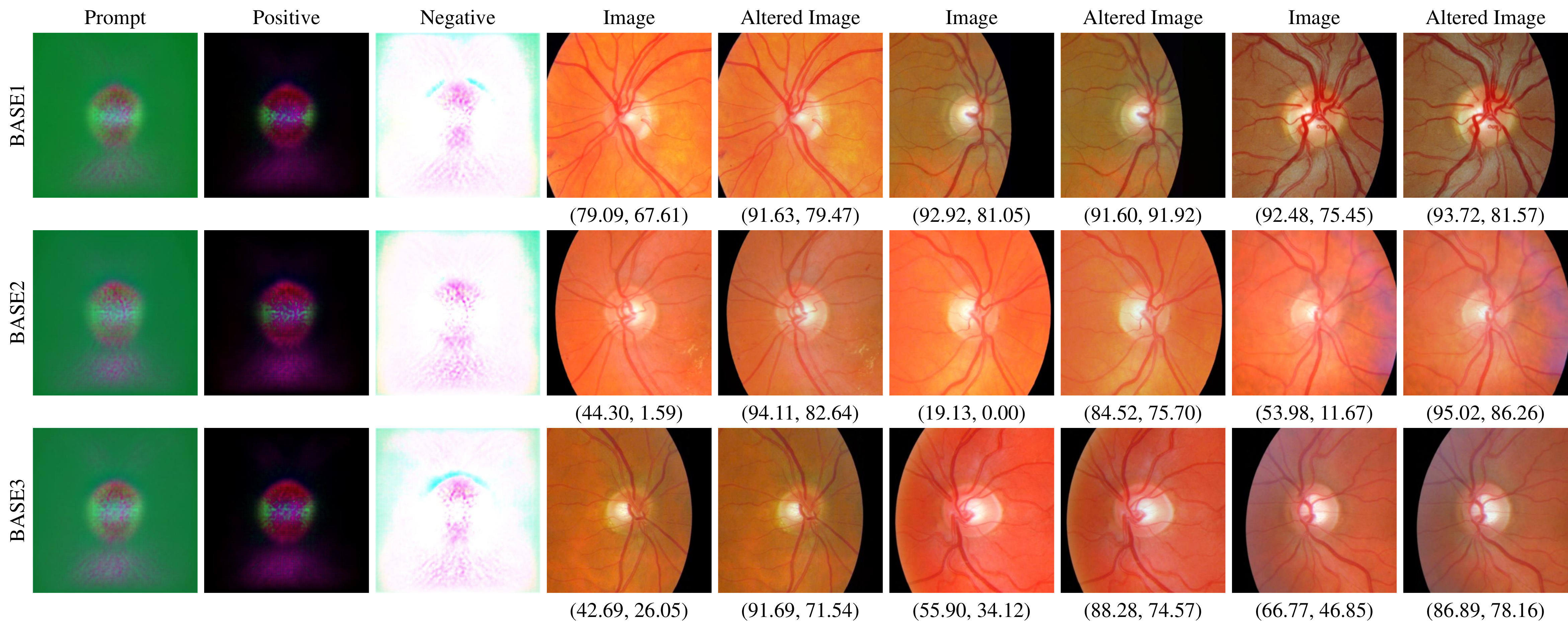}
  \caption{
Visualization of the estimated prompt, its positive and negative components on three target domains of joint OC/OD segmentation. We channel-wise re-scale the range of prompt to [0, 255] for better visualization. We also display three groups of target domain images and their counterparts altered with prompt from three target domains, respectively. The performance ($\mathcal{D}_{OD}$(\%), $\mathcal{D}_{OC}$(\%)) of OC/OD segmentation of the pre-trained source model on target domain images and their counterparts altered with prompt are also displayed for reference.
  }
  \label{fig:prompt}
\end{figure*}

We visualize the estimated domain-aware prompts on three target domains of joint OC/OD segmentation task, as shown in~\figurename{~\ref{fig:prompt}}.
It shows that the estimated prompt is domain-specific.
We also display three groups of target domain images and their counterparts altered with prompt from three target domains, as shown in the fourth column to the ninth column of~\figurename{~\ref{fig:prompt}}.
The segmentation performance of the pre-trained source model on these images are also displayed for reference.
It reveals minor changes (even human-imperceptible) are modified by the prompt.
Even though, the segmentation performance on the target domain images and their altered counterpart by the pre-trained source model with the same group of parameters can be substantially improved, especially on the target domain `BASE2' with relative large domain discrepancy.
We believe the substantially improved performance can be attributed to the fact that the estimated domain-aware prompt effectively aligns BN statistics in the source model and the updated source model which takes the altered target image as input.

\subsection{Analysis of Performance Gain}
\begin{figure}[]
  \centering
  \includegraphics[width=0.5\textwidth]{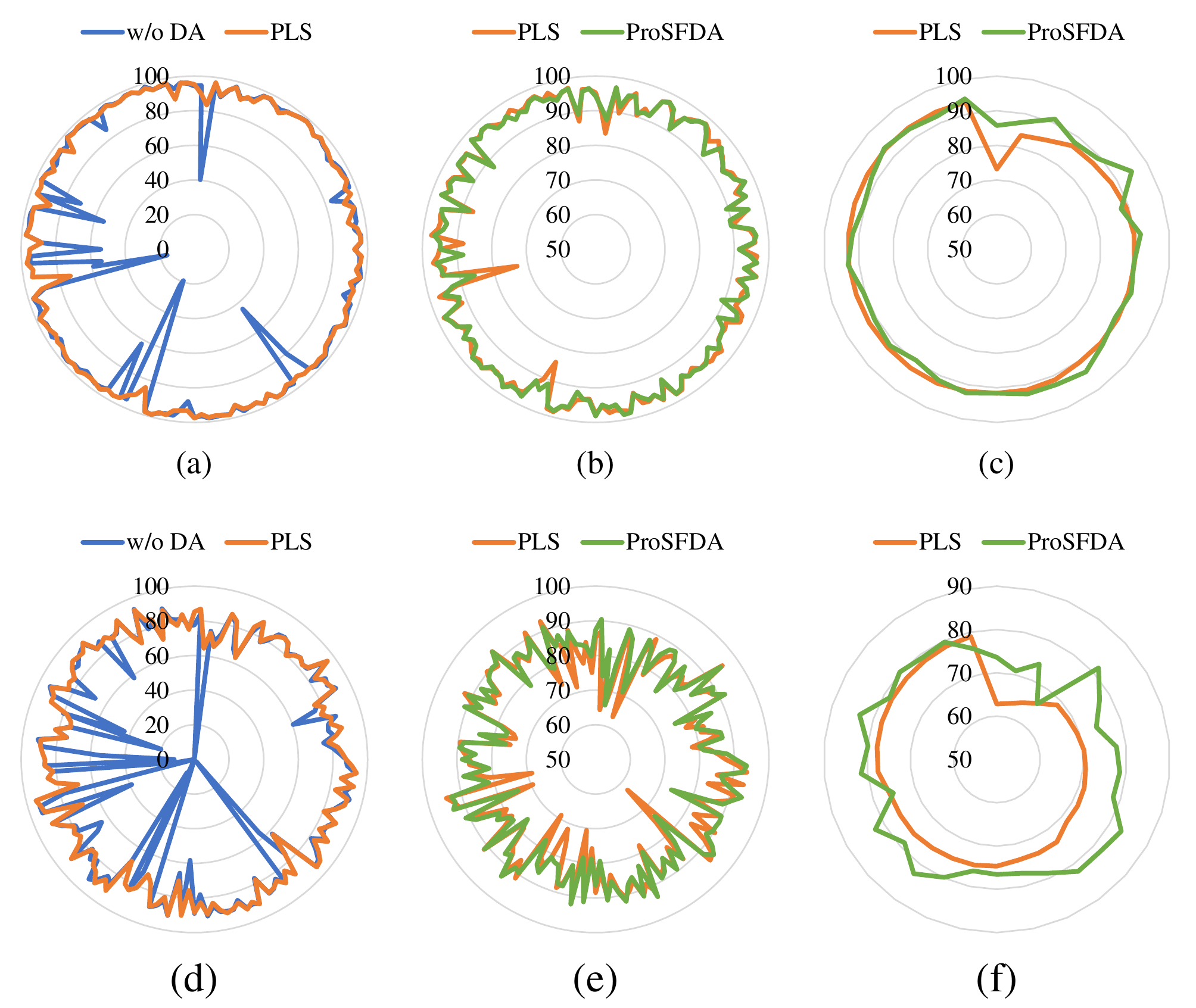}
  \caption{
The performance (\%) of (a) and (d): the source model on target domain images (\ie, w/o DA) and altered target domain images (\ie, PLS), (b) and (e): PLS and our ProSFDA, and (c) and (f): PLS and our ProSFDA on 20\% altered images of PLS's worst segmentation. (a), (b), and (c) shows the performance for OD segmentation on `BASE2', whereas (d), (e), and (f) displays the performance for OC segmentation on `BASE2'.
  }
  \label{fig:gain}
\end{figure}

To further investigate PLS and FAS, we compare the performance of `w/o DA', PLS, and `PLS+FAS' (\ie, ProSFDA) on all images of `BASE2', as shown in~\figurename{~\ref{fig:gain}} (a), (b), (d), and (e).
We also display the performance of PLS and ProSFDA on the altered images whose performance of PLS are the worst 20\%, as shown in~\figurename{~\ref{fig:gain}} (c) and (f).
Their low segmentation performance can be attributed to that these altered images are still far away from the prototype of the left images in the feature space. 
\figurename{~\ref{fig:gain}} (a) and (d) shows that PLS can improve the overall segmentation performance on both OD and OC segmentation, especially on the target domain images with worse performance, confirming the effectiveness of the estimated domain-aware prompt.
\figurename{~\ref{fig:gain}} (b) and (e) reveals that FAS further boosts the segmentation performance on the altered target domain images.
Moreover, \figurename{~\ref{fig:gain}} (c) and (f) demonstrate that FAS can improve the performance on the altered target domain images with worse performance.
We believe the increased performance on these altered images can be attributed to that FAS can amend the source model toward extracting compact features.